\documentclass{article}
\usepackage{spconf}
\usepackage{amsmath}
\usepackage{amssymb}
\usepackage{epsfig}
\usepackage{multirow}
\usepackage{multicol}
\usepackage{booktabs}
\usepackage{gensymb}
\usepackage{makecell}
\usepackage{pgfplots}
\pgfplotsset{compat=newest}
\usepgfplotslibrary{groupplots}
\usepgfplotslibrary{dateplot}
\usepackage{booktabs}
\usepackage{tikzscale}
\usepackage{hyperref}
\hypersetup{hidelinks,
	colorlinks=true,
	allcolors=black,
	pdfstartview=Fit,
	breaklinks=true}
\usepackage{algorithmic}
\usepackage{array}
\usepackage{url}

\title{Segment Change Model (SCM) for Unsupervised Change detection in VHR Remote Sensing Images: a Case Study of Buildings}

\name{Xiaoliang Tan, Guanzhou Chen\sthanks{Corresponding author: cgz@whu.edu.cn; zxdlmars@whu.edu.cn}, Tong Wang, Jiaqi Wang, Xiaodong Zhang\footnotemark[1]} 
\address{State Key Laboratory of Information Engineering in Surveying, Mapping and Remote Sensing,\\ 
Wuhan University, Wuhan 430079, China}

\begin{document}
%
\maketitle

\begin{abstract}
The field of Remote Sensing (RS) widely employs Change Detection (CD) on very-high-resolution (VHR) images. A majority of extant deep-learning-based methods hinge on annotated samples to complete the CD process. Recently, the emergence of Vision Foundation Model (VFM) enables zero-shot predictions in particular vision tasks. In this work, we propose an unsupervised CD method named Segment Change Model (SCM), built upon the Segment Anything Model (SAM) and Contrastive Language-Image Pre-training (CLIP). Our method recalibrates features extracted at different scales and integrates them in a top-down manner to enhance discriminative change edges. We further design an innovative Piecewise Semantic Attention (PSA) scheme, which can offer semantic representation without training, thereby minimize pseudo change phenomenon. Through conducting experiments on two public datasets, the proposed SCM increases the mIoU from 46.09\% to 53.67\% on the LEVIR-CD dataset, and from 47.56\% to 52.14\% on the WHU-CD dataset. Our codes are available at: \href{https://github.com/StephenApX/UCD-SCM}{https://github.com/StephenApX/UCD-SCM}.
\end{abstract}


\section{Introduction}
\label{sec_intro}
Remote sensing (RS) change detection (CD) aims to analyze change regions from two or more corresponding images taken at different times. As a fundamental task in RS community, change detection on very-high-resolution (VHR) images, propels a variety applications for disaster evaluation, land-use and land-cover (LULC) investigation and urban planning \cite{tang_unsupervised_2022}. 

Unsupervised change detection (UCD) methods are usually deployed under limited samples. Existing UCD methods can be divided into traditional approaches and deep-learning (DL) approaches. Traditional UCD methods attempt to acquire a difference map at various levels, such as image-level, feature-level and post-classification-level \cite{turgay_celik_unsupervised_2009}. Some DL-based methods adopt pre-trained convolutional neural network (CNN) as a feature extractor in order to better capture difference at different feature levels \cite{jiang_convolutional_2016, saha_unsupervised_2019}. Others excavate reliable information from images as supervisory signals, and conduct training to distinguish changed or unchanged areas. For instance, Du \textit{et al.} proposed a Deep Slow Feature Analysis (DSFA) to find unchanged pixels as training samples and generate change map after optimization \cite{du_unsupervised_2019}. GMCD consists of a pseudo label generation mechanism based on metric learning to accomplish model training \cite{tang_unsupervised_2022}. Despite the efficiency of DL in generating valid change maps, existing methods disregard the characteristic and inner correlation among different scales. Besides, other training-based methods rely on limited information, which may result in the scarcity of semantic representation and phenomena of pseudo-change.

Recently, vision foundation models (VFMs) and multimodal large language models (MLLMs) have shown great potentials in computer vision and natural language processing fields. Under zero-shot conditions, VFMs such as Segment Anything Model (SAM) \cite{kirillov_segment_2023} and FastSAM \cite{zhao_fast_2023} are capable of recognizing visual objects and generate fine-grained masks. Contrastive Language-Image Pre-training (CLIP) model \cite{radford_learning_2021} can process inputs from different modalities and generate comparable embeddings, establishing connections between text and image. Combining SAM with CLIP, we can simultaneously obtain spatial understanding for segmentation and semantic understanding for classification. However, gaps still exist between open vocabulary and task-specific category, making it unfit to UCD task.
%

In this work, we propose a Segment Change Model (SCM) for unsupervised change detection to address the aforementioned issues. Multi-scale features from bi-temporal images are firstly extracted from FastSAM, then fed into a Recalibrated Feature Fusion (RFF) module in order to better reveal local meanwhile global changes and preserve distinct change edges. Cooperating with SAM and CLIP models, we innovatively design a Piecewise Semantic Attention (PSA) scheme, which can introduce semantic understanding through simple texts to filter out pseudo changes. Afterwards, pixel-wise cosine distance is computed between concatenated features, then segmented into a binary change map through a global OTSU threshold algorithm. In this way, our proposed method enhances integration of multi-scale features from RS images, reduces pseudo change occurrence and improves performances over existing methods.

\section{Methodology}
\label{sec_method}

\begin{figure*}[htbp]
	\centering
  \includegraphics[width=1\linewidth]{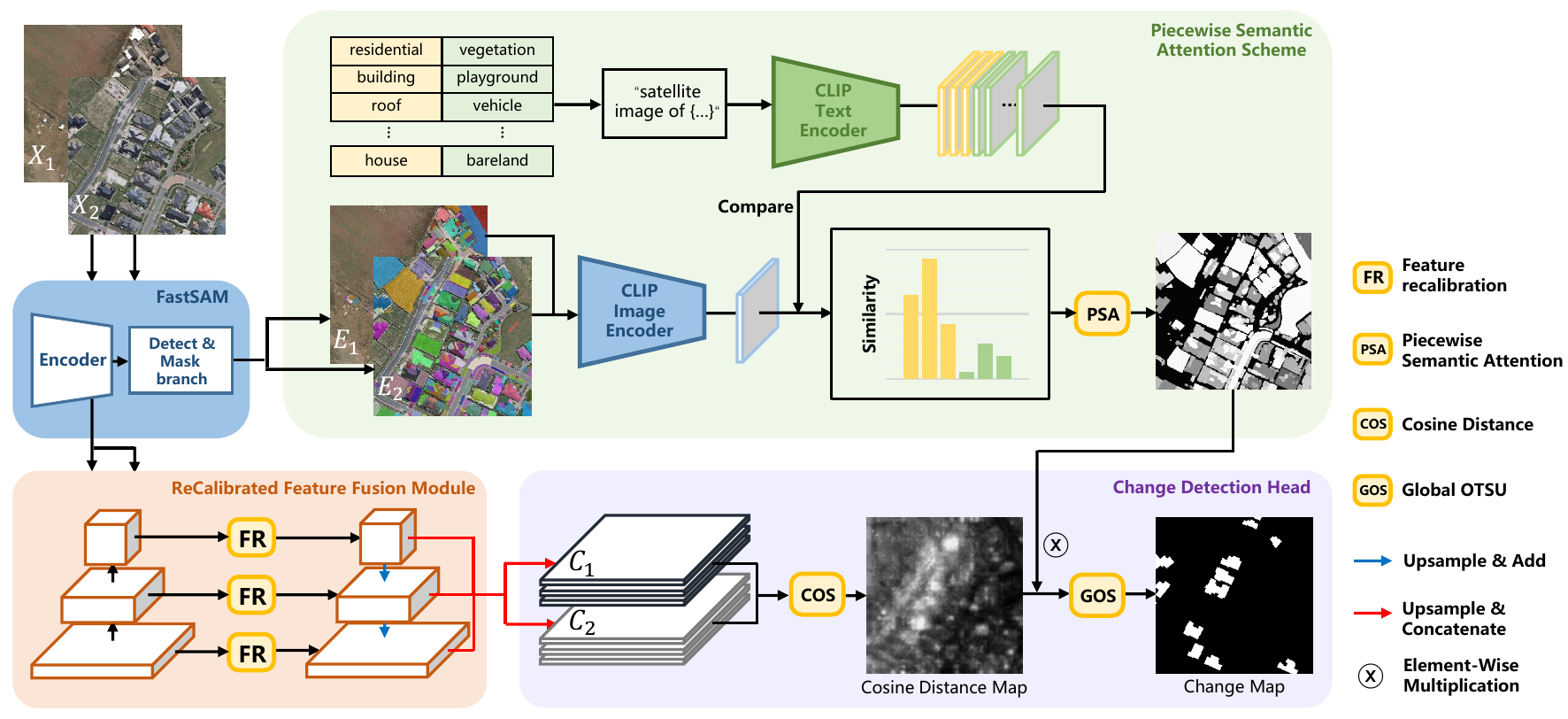}
  \caption{Framework of Segment Change Model (SCM).}
  \label{fig_method}
\end{figure*}

\subsection{Overview}
\label{ssec_overview}
The overall SCM framework is shown in Fig.\ref{fig_method}. The flowchart of the proposed SCM framework generally starts with feeding a pair of bi-temporal RS images $\mathbf{X_{1}} \in \mathbb{R}^{H \times W \times C}$ and $\mathbf{X_{2}} \in \mathbb{R}^{H \times W \times C}$ into a pre-trained FastSAM. The main architecture of FastSAM consists of a CNN-based encoder, a detect branch and a mask branch. The encoder can be divided into five stages, where we obtain the feature maps with different spatial and channel dimensions from last three stages, denoted as $\left\{ \mathbf{F_{3}}, \mathbf{F_{4}}, \mathbf{F_{5}} \right\}$. We upsample and fuse them into two concatenations $\left\{ \mathbf{C_{1}}, \mathbf{C_{2}} \right\}$ with same spatial size as the input image through Recalibrated Feature Fusion (RFF) module presented in Sec.\ref{ssec_RFF}. A difference map between $\mathbf{C_{1}}$ and $\mathbf{C_{2}}$ is calculated with cosine similarity at channel-wise axis, which depicts pixel-wise similarity between $\mathbf{X_{1}}$ and $\mathbf{X_{2}}$. This process can be formulated as:
\begin{equation}
\setlength{\abovedisplayskip}{4pt}
\begin{aligned}
  {diff}_{(i,j)} &= 1 - \frac{C_1 \cdot C_2}{\left\| C_1 \right\|_2 \cdot \left\| C_2 \right\|_2} \\ 
     &= 1 - \frac{\sum_{c=0}^{k} C_1^{(i,j,c)} C_2^{(i,j,c)}}{\sqrt{\sum_{c=0}^{k} (C_1^{(i,j,c)})^2} \sqrt{\sum_{c=0}^{k} (C_2^{(i,j,c)})^2}}, \\
     &i \in [1,H], j \in [1,W].
\end{aligned}
\setlength{\belowdisplayskip}{4pt}
\end{equation}
where $i,j$ respectively depict the pixel location of image's height and width. In Sec.\ref{ssec_PSA}, we generate a Piecewise Semantic Attention (PSA) map and multiply with the difference map to filter out pseudo change phenomenon. Afterwards, a global OTSU thresholding algorithm is conducted on all non-zero values to search an optimal threshold for discriminating change and non-change area.

\subsection{Recalibrated Feature Fusion Module}
\label{ssec_RFF}
Although VFMs excel in extracting discriminating feature representation at various scales, simple concatenation of multi-scale feature maps may neglect their correlations and lead to imblance when computing their difference map under UCD circumstance. We construct a Recalibrated Feature Fusion (RFF) module in order to integrate low-level local features with high-level global features and restore the semantic correlations across different scales in parameter-free manner.

We fisrt obtain a set of feature maps $\left\{ \mathbf{F_{3}}, \mathbf{F_{4}}, \mathbf{F_{5}} \right\}$ from last three stages of encoder of the FastSAM. At each scale, we recalibrate each feature map with a size of $H \times W \times C$ by calculating mean value of each channel and generating the weight sequence with a size of $1 \times 1 \times C$. By multiplying feature map and its weight sequence, we attempt to model inter-dependencies across different channels. Then, recalibrated features are integrated in a top-down manner. Starting from the last stage, feature maps from higher level are sampled equidistantly along channel dimension and interpolated along spatial dimension to align with feature maps from lower level. In this way, the resampled feature map from higher level is then merged with corresponding low level map by element-wise addition, aiming to semantically enhance lower level maps meanwhile maintain their local activations. Afterwards, we spatially upsample feature map with same spatial size as the input image and concatenate them into $\left\{ \mathbf{C_{1}}, \mathbf{C_{2}} \right\}$. 

\subsection{Piecewise Semantic Attention}
\label{ssec_PSA}

In order to filter out pseudo change phenomenon, which is mainly caused by lack of semantic understanding in UCD task, we design a scheme based on SAM and CLIP and generate a Piecewise Semantic Attention (PSA) map.

FastSAM is firstly utilized to acquire every possible objects' segmentation masks $\left\{ \mathbf{E_{1,2}}\right\}$from an input image. In our scheme, we regard building change as the principal change target. In order to distinguish each segmented image patch between building and non-building classes, we construct two groups of texts which separately represents building related and non-building related objects:

\begin{table}[htbp]
  \centering
  \begin{tabular}{p{3.3cm}<{\centering}|p{4.4cm}<{\centering}}
    \toprule
    building & non-building \\
    \midrule
    roof, rooftop, building, house, apartment, residential, factory & baseball, diamond, bareland, swimming pool, basketball court,  roundabout, playground \\
    \bottomrule
  \end{tabular}%
\end{table}%
Afterwards, segmented image patches and predetermined groups of texts are respectively sent into the image encoder and text encoder of CLIP to acquire comparable embeddings. For each image patch, cosine similarity between image embedding and text embeddings are calculated and fed into the softmax function to generate categorical distributions. We summarize them into building (bld) class and non-building class probability:
\begin{equation}
\setlength{\abovedisplayskip}{4pt}
\begin{split}
  P_{bld} &= \sum\nolimits_{c \in \{bld \ related \ cls\}}^{\ } p_c \\
\end{split}
\end{equation}
After iterating each image patch and acquiring its building class probability, we initialize a zero score map with same size as input image and assign each pixel with every segmented mask and its corresponding probability. Moreover, in order to integrate bi-temporal information and avoid possible misclassification by CLIP classifier, we conduct an element-wise addition between two score maps from $\left\{ \mathbf{E_{1}}, \mathbf{E_{2}} \right\}$, and generate a semantic attention map through a piecewise remapping function:
\begin{equation}
\setlength{\abovedisplayskip}{4pt}
  PSA = \left\{\begin{aligned}
  1,\  & when \ P_{bld} >= 0.5; \\
  P_{bld} * 2 ,\ & when \ 0 < P_{bld} < 0.5; \\
  0,\ & background. \end{aligned} \right.
\end{equation}
which is used to multiply with the difference map to filter out non-building changes.


\section{Experiments}
\label{sec_exp}
\begin{figure*}[htb]
\flushleft
\centering
\begin{minipage}[b]{0.11\linewidth}
  \centering{\epsfig{figure=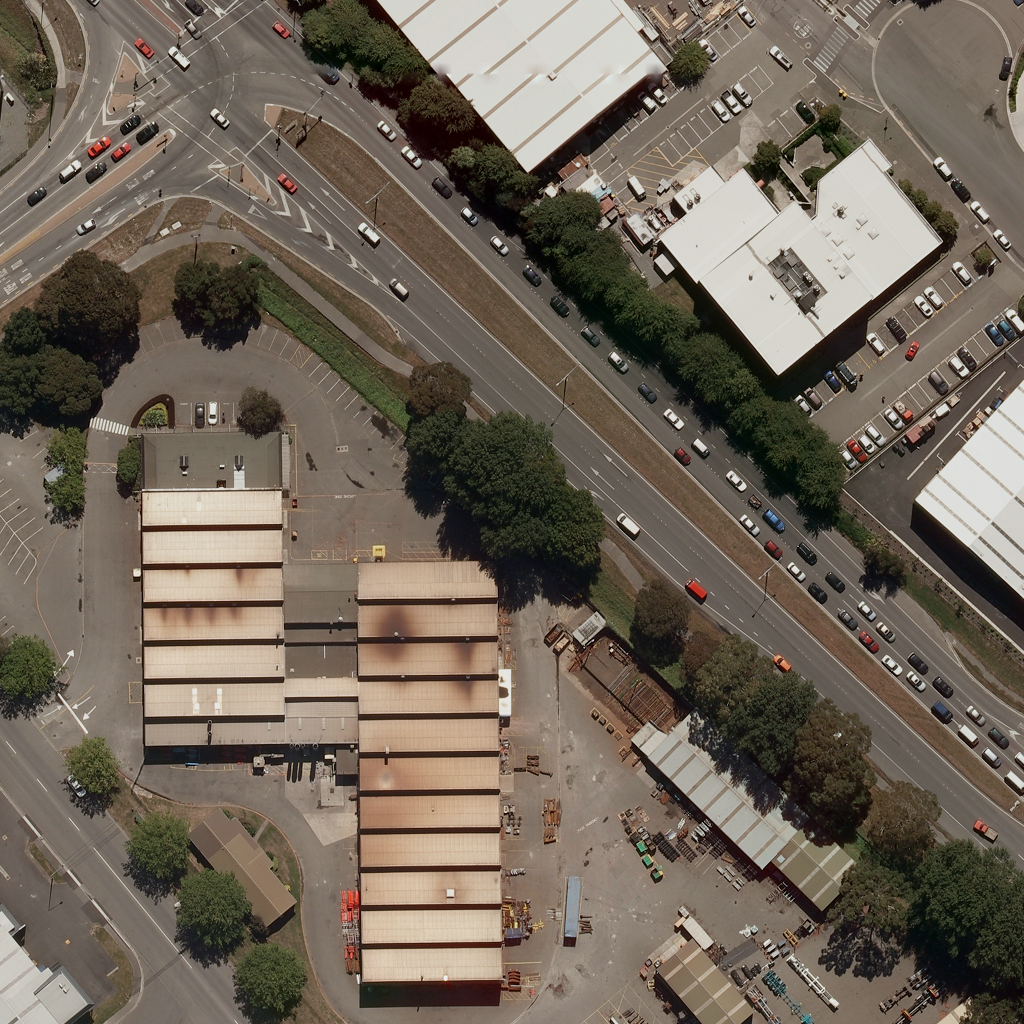,width=1.7cm}} 
\end{minipage}\hspace{-0.2cm}
\begin{minipage}[b]{0.11\linewidth}
  \centering{\epsfig{figure=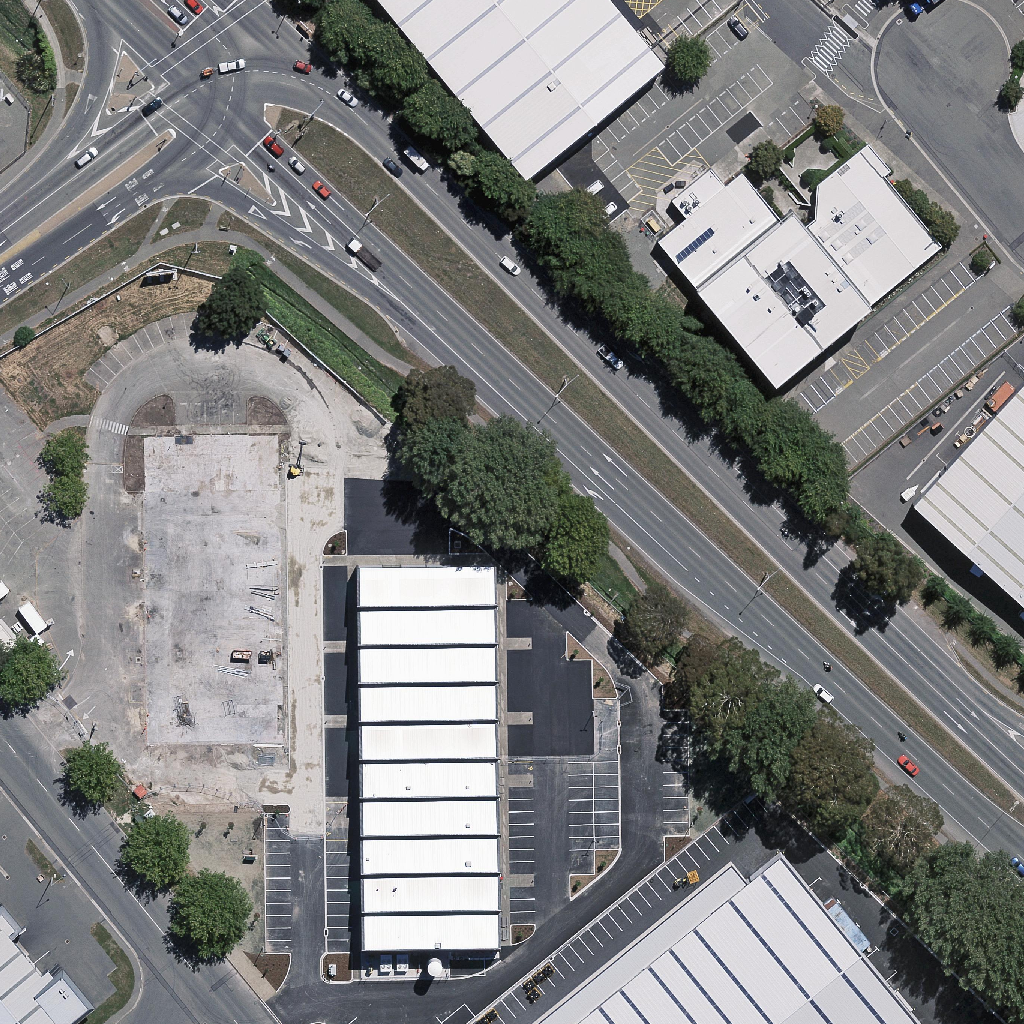,width=1.7cm}} 
\end{minipage}\hspace{-0.2cm}
\begin{minipage}[b]{0.11\linewidth}
  \centering{\epsfig{figure=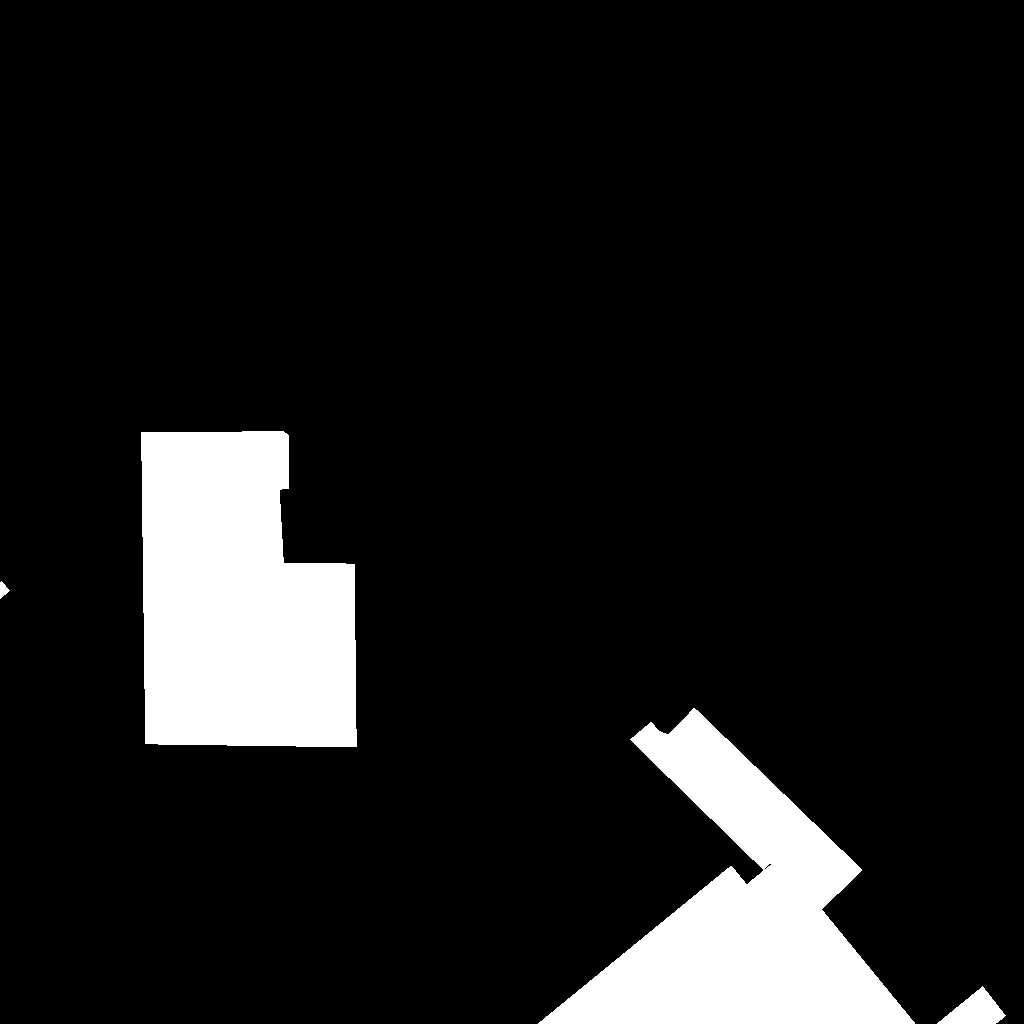,width=1.7cm}} 
\end{minipage}\hspace{-0.2cm}
\begin{minipage}[b]{0.11\linewidth}
  \centering{\epsfig{figure=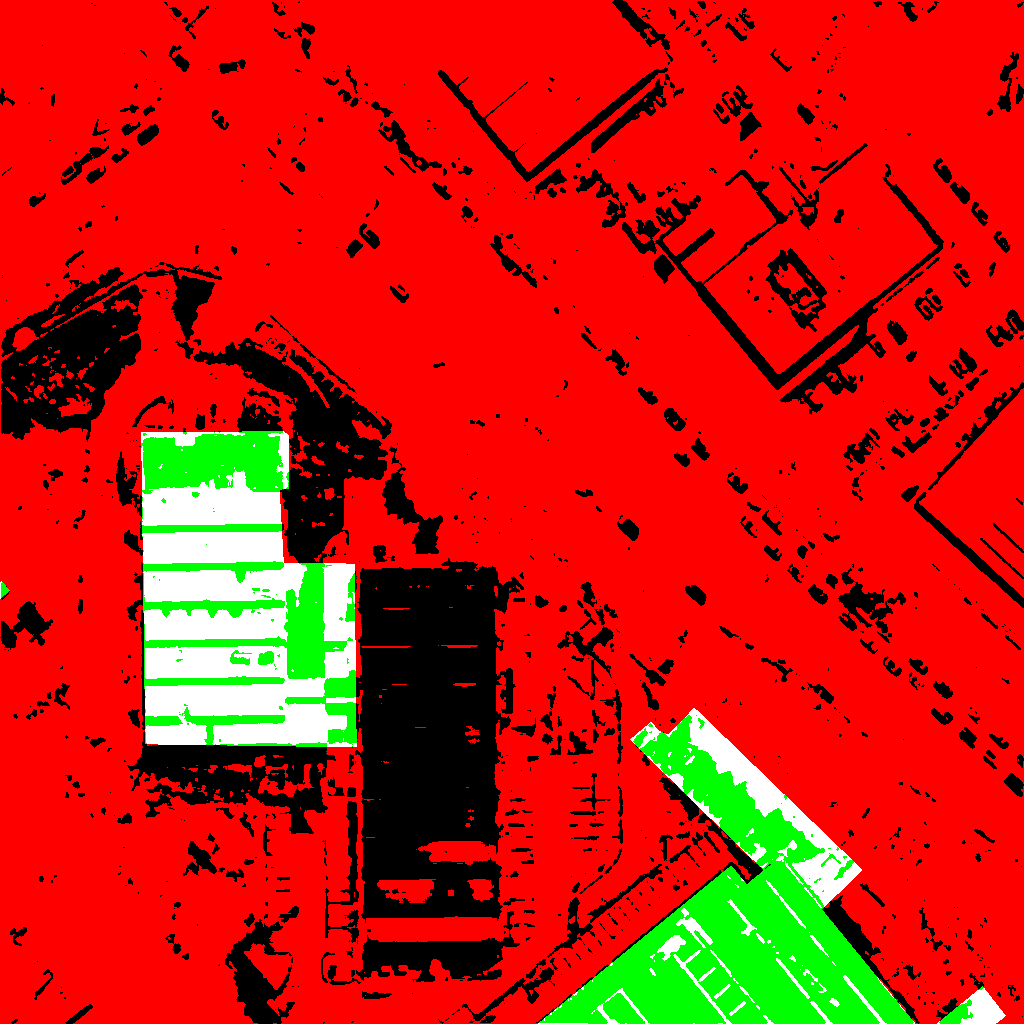,width=1.7cm}} 
\end{minipage}\hspace{-0.2cm}
\begin{minipage}[b]{0.11\linewidth}
  \centering{\epsfig{figure=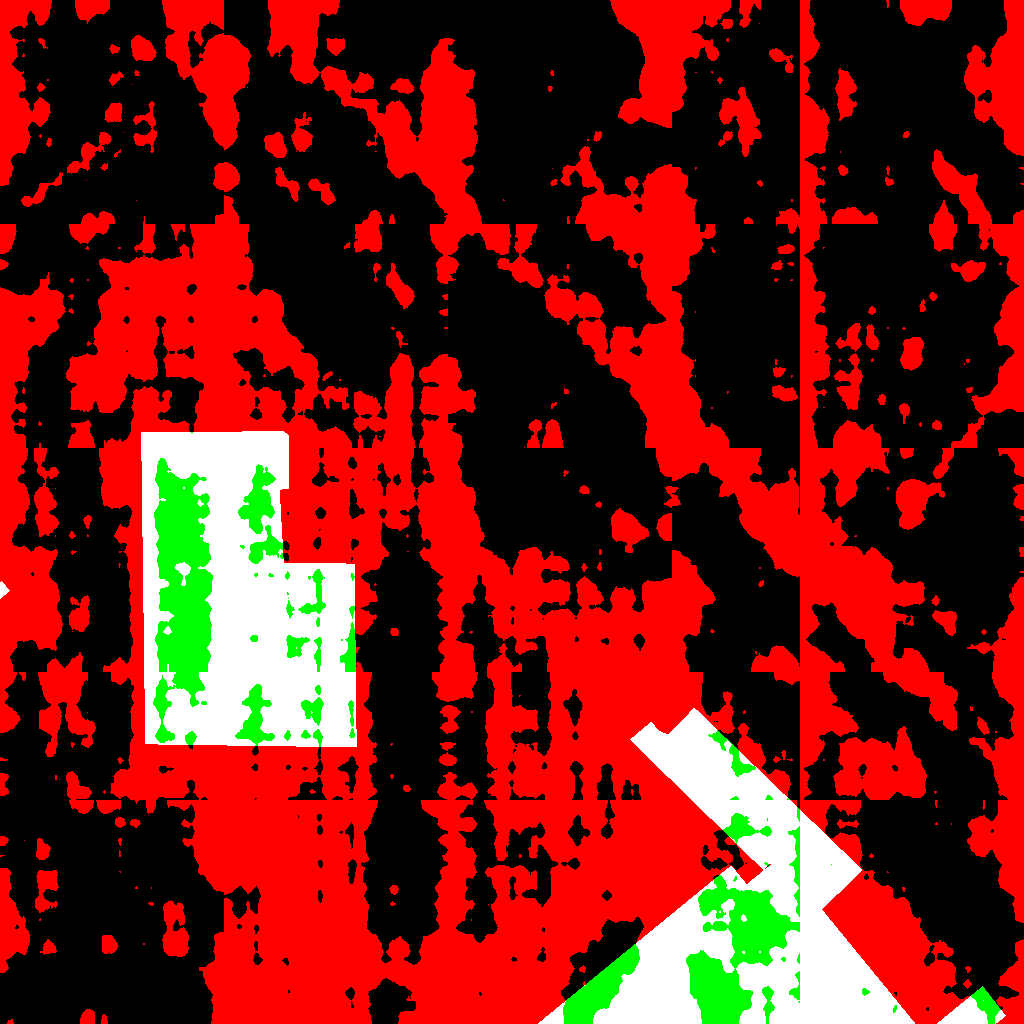,width=1.7cm}} 
\end{minipage}\hspace{-0.2cm}
\begin{minipage}[b]{0.11\linewidth}
  \centering{\epsfig{figure=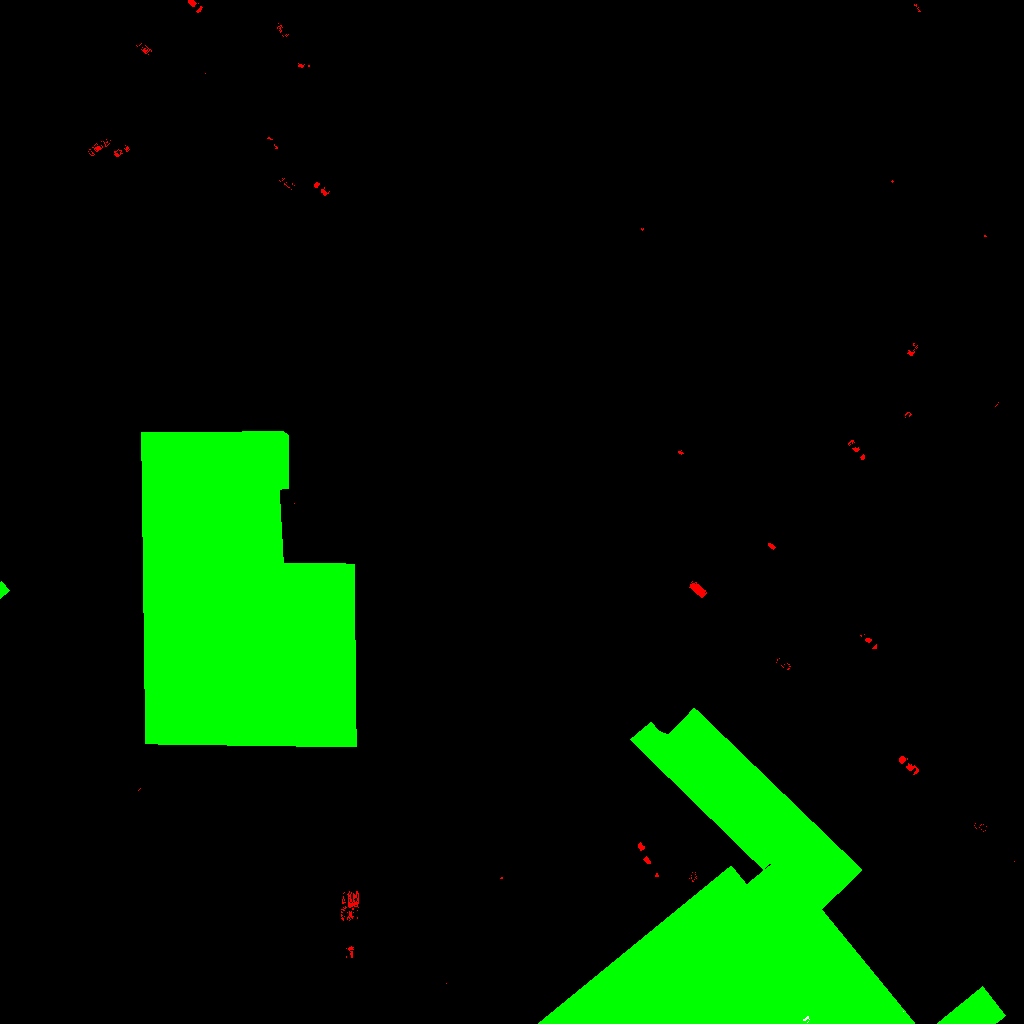,width=1.7cm}} 
\end{minipage}\hspace{-0.2cm}
\begin{minipage}[b]{0.11\linewidth}
  \centering{\epsfig{figure=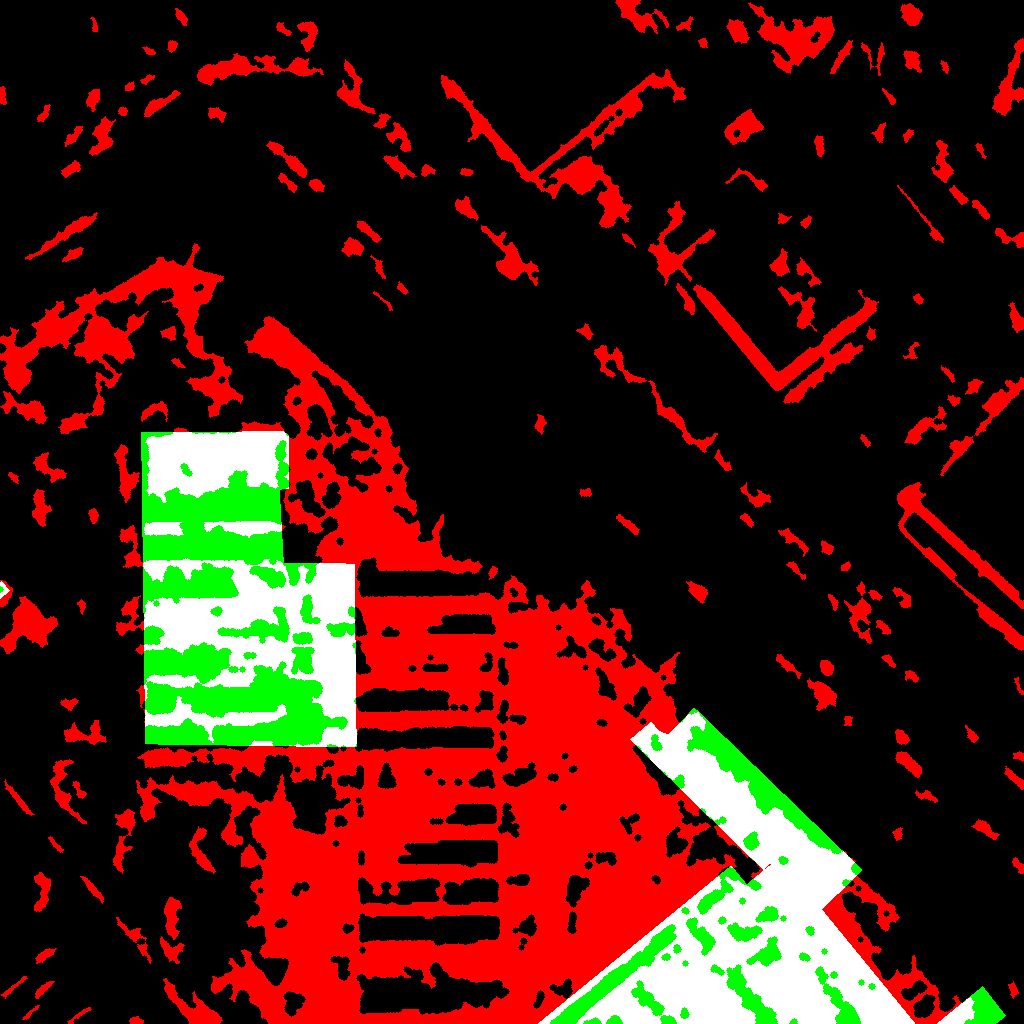,width=1.7cm}} 
\end{minipage}\hspace{-0.2cm}
\begin{minipage}[b]{0.11\linewidth}
  \centering{\epsfig{figure=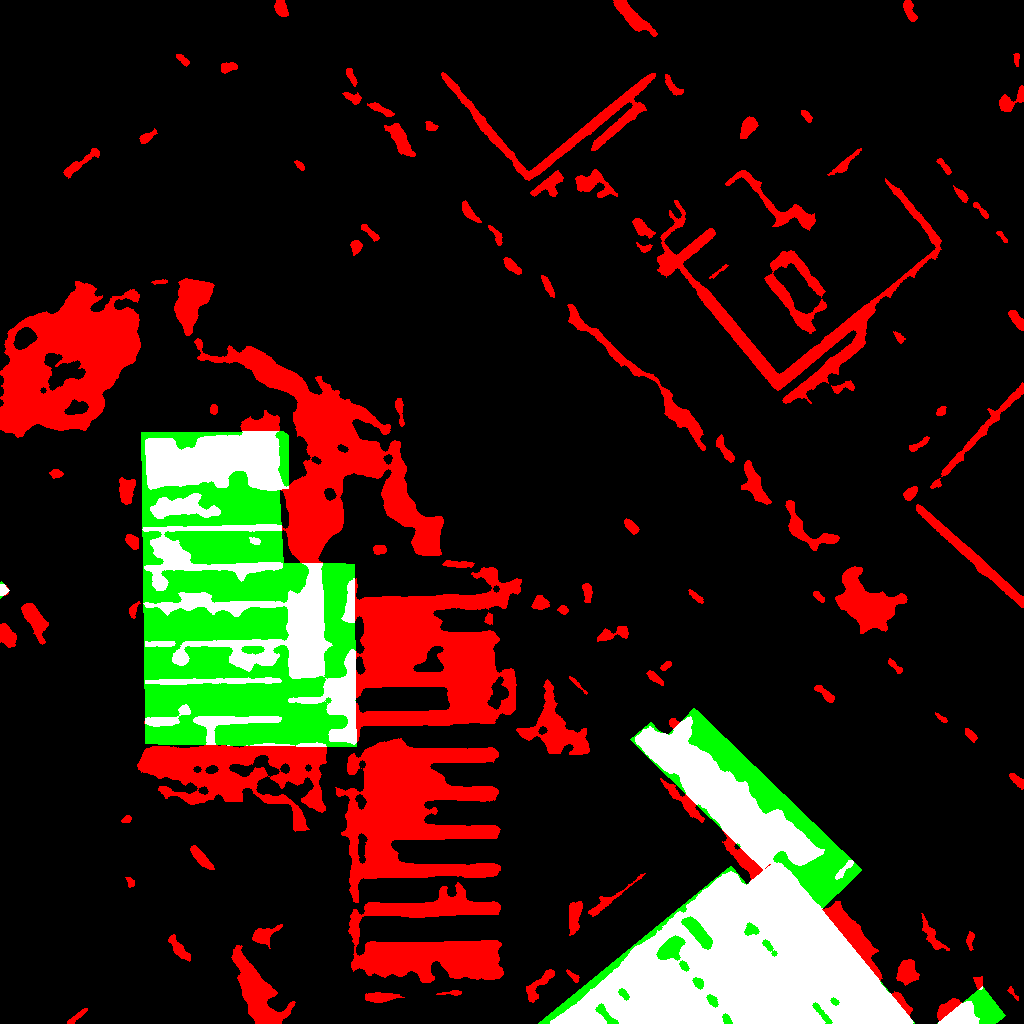,width=1.7cm}} 
\end{minipage}\hspace{-0.2cm}
\begin{minipage}[b]{0.11\linewidth}
  \centering{\epsfig{figure=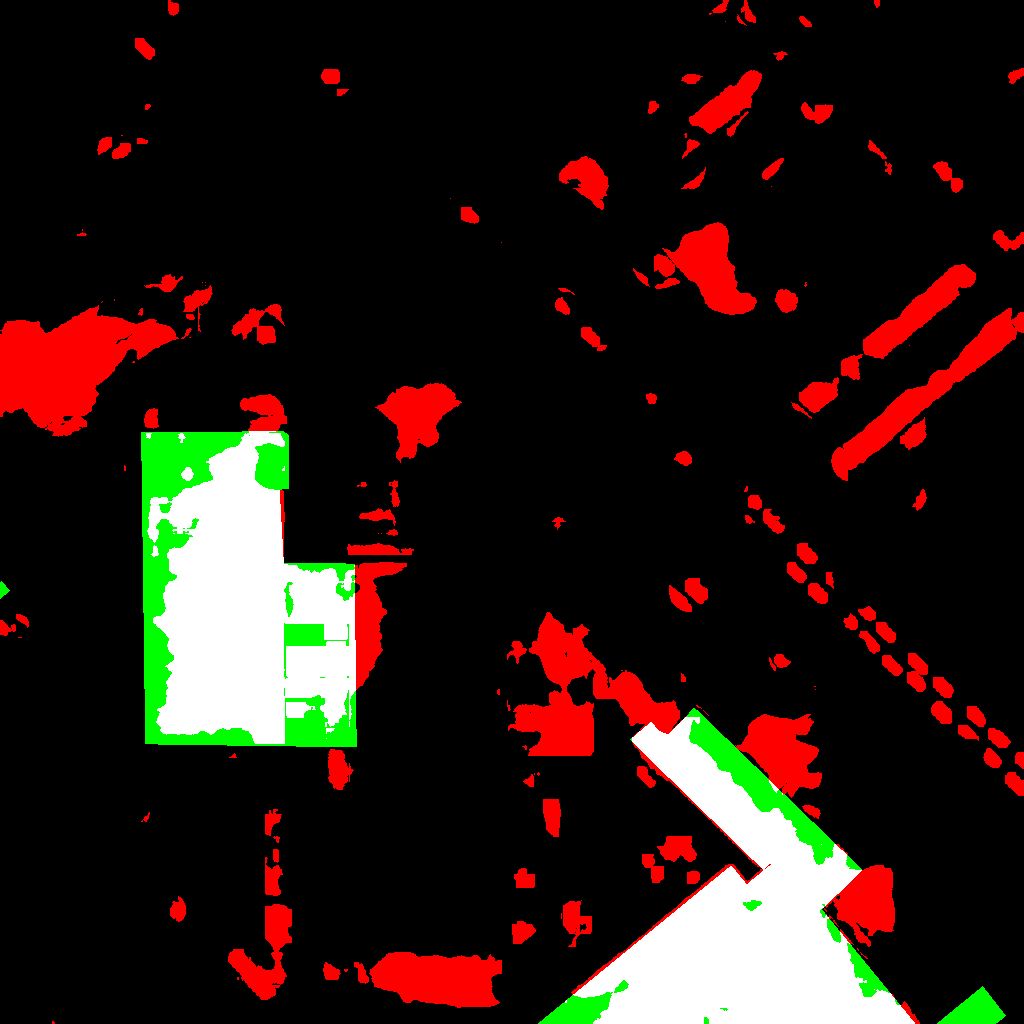,width=1.7cm}} 
\end{minipage}\hspace{-0.2cm}
\vfill
\vspace{0.1cm}
\begin{minipage}[b]{0.11\linewidth}
  \centering{\epsfig{figure=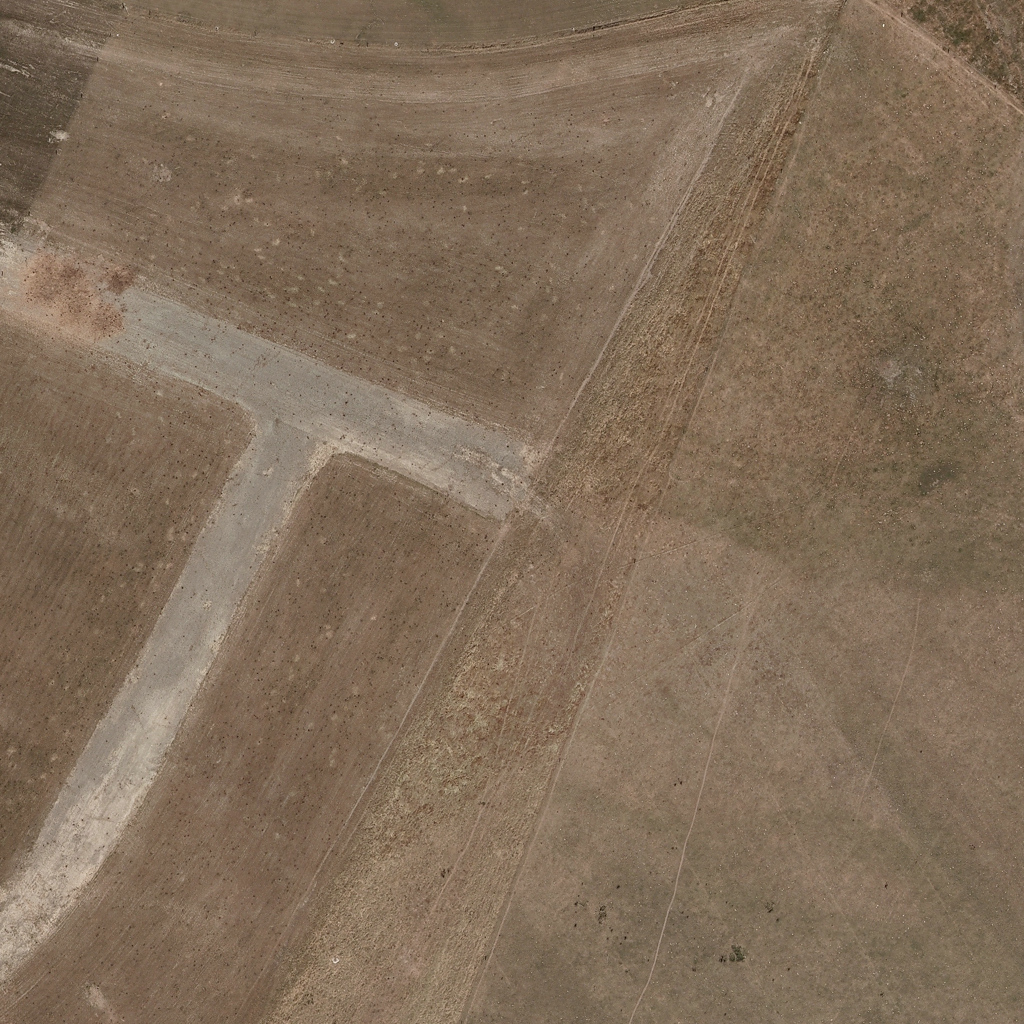,width=1.7cm}} 
  \centerline{(a)}\medskip
\end{minipage}\hspace{-0.2cm}
\begin{minipage}[b]{0.11\linewidth}
  \centering{\epsfig{figure=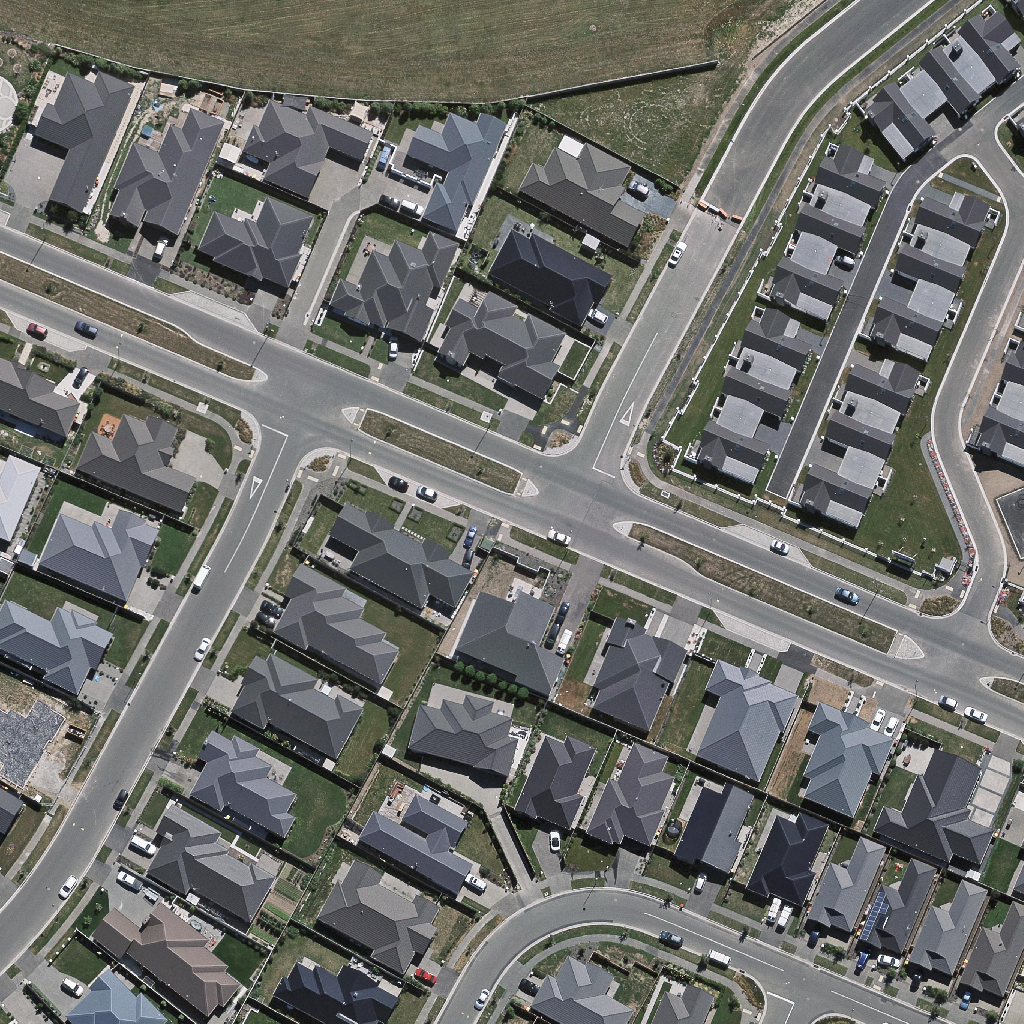,width=1.7cm}} 
  \centerline{(b)}\medskip
\end{minipage}\hspace{-0.2cm}
\begin{minipage}[b]{0.11\linewidth}
  \centering{\epsfig{figure=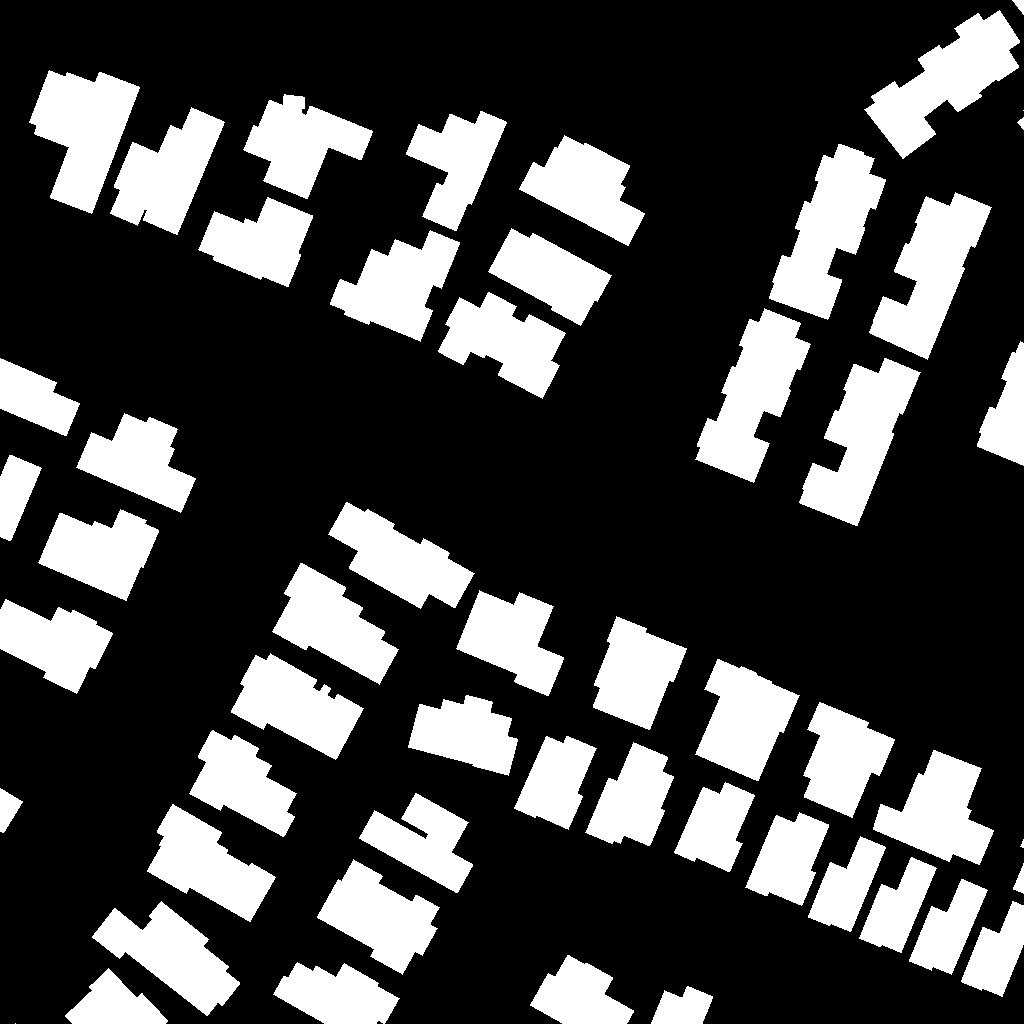,width=1.7cm}} 
  \centerline{(c)}\medskip
\end{minipage}\hspace{-0.2cm}
\begin{minipage}[b]{0.11\linewidth}
  \centering{\epsfig{figure=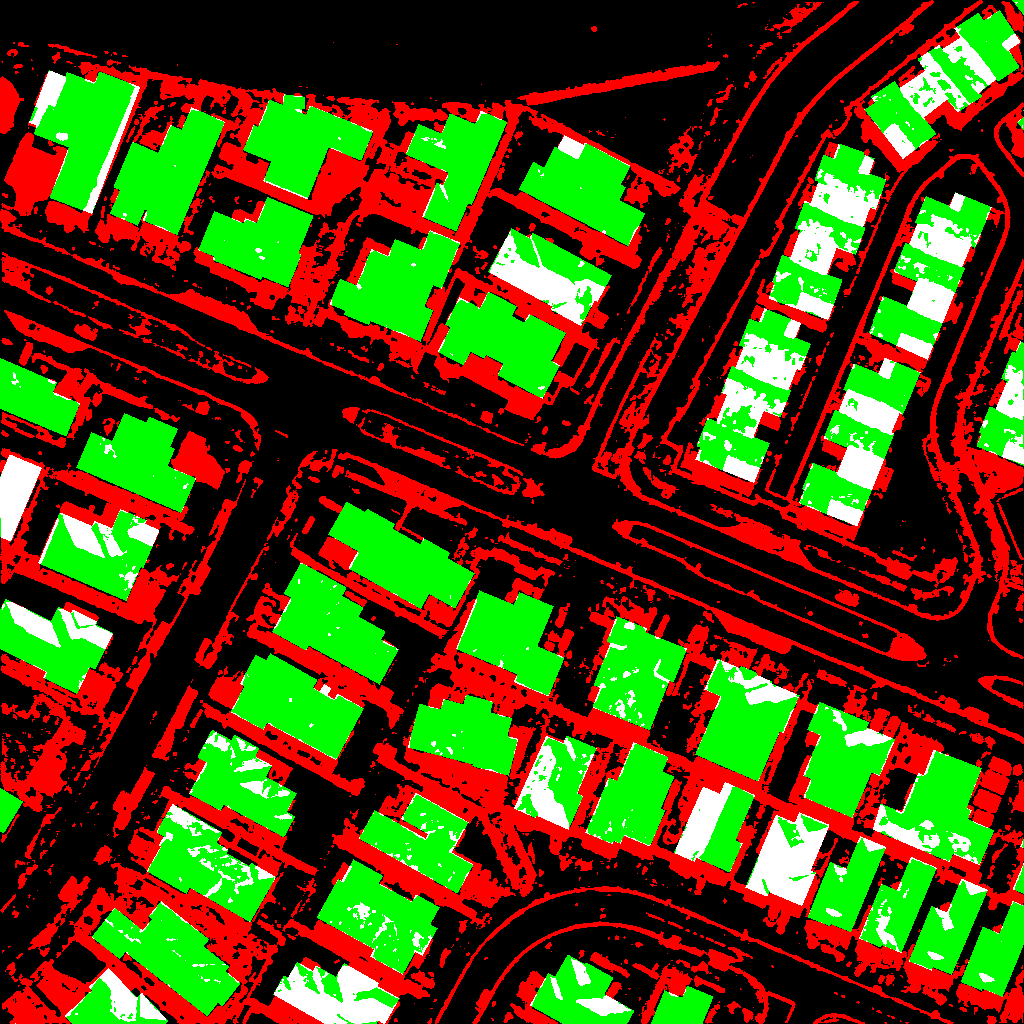,width=1.7cm}} 
  \centerline{(d)}\medskip
\end{minipage}\hspace{-0.2cm}
\begin{minipage}[b]{0.11\linewidth}
  \centering{\epsfig{figure=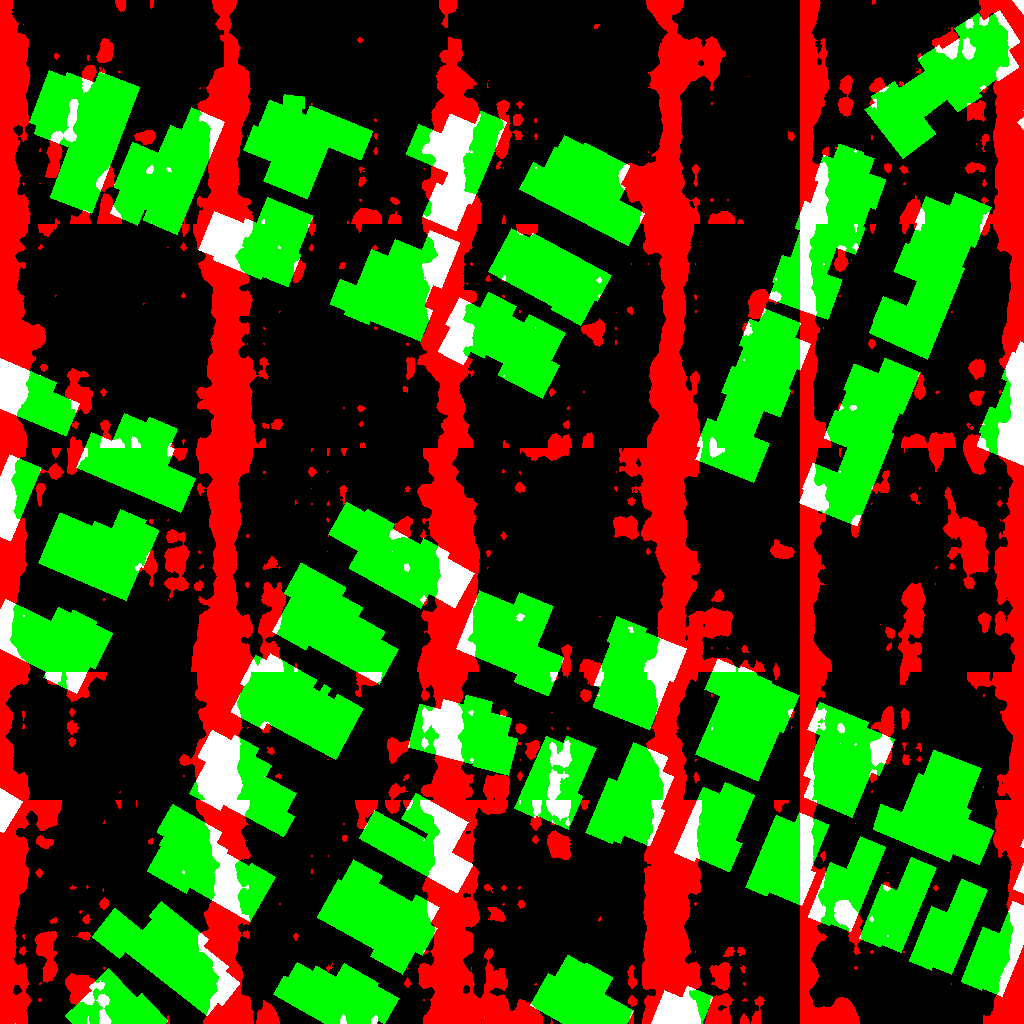,width=1.7cm}} 
  \centerline{(e)}\medskip
\end{minipage}\hspace{-0.2cm}
\begin{minipage}[b]{0.11\linewidth}
  \centering{\epsfig{figure=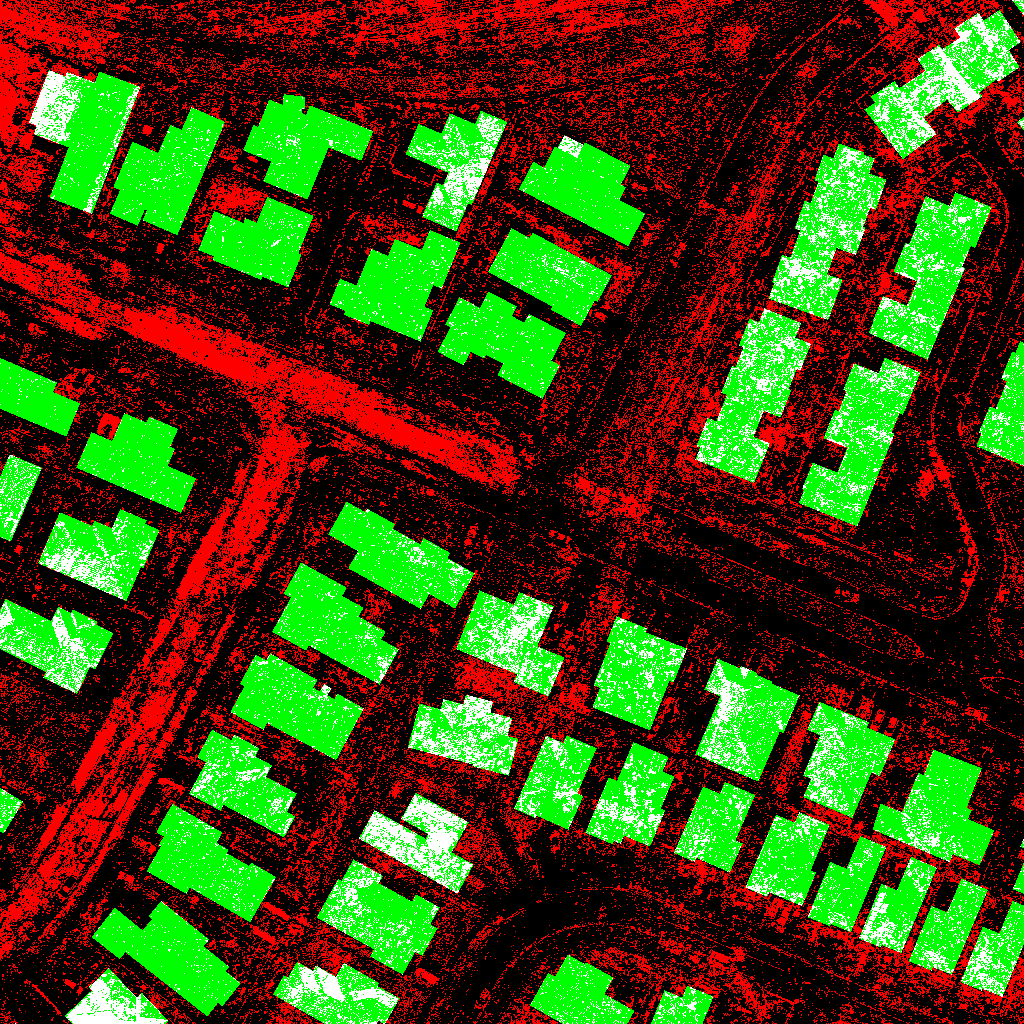,width=1.7cm}} 
  \centerline{(f)}\medskip
\end{minipage}\hspace{-0.2cm}
\begin{minipage}[b]{0.11\linewidth}
  \centering{\epsfig{figure=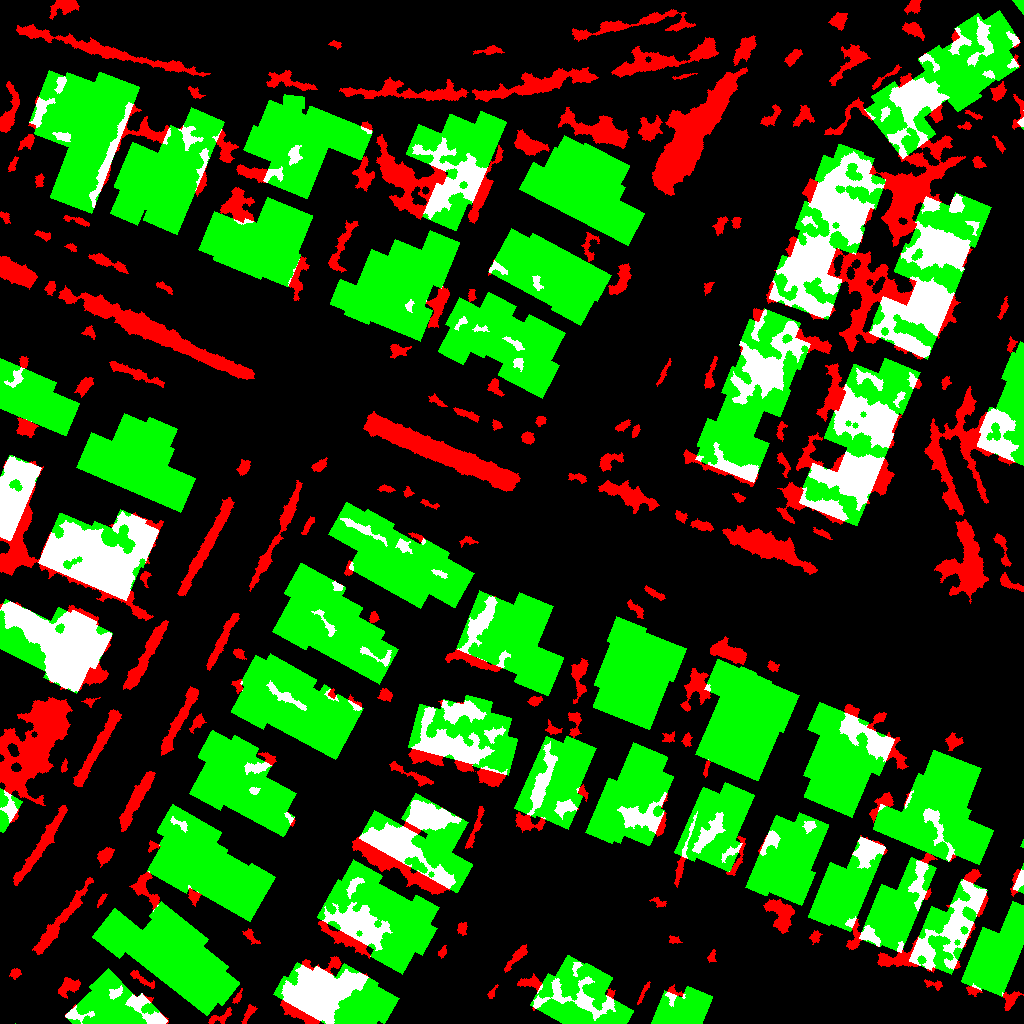,width=1.7cm}} 
  \centerline{(g)}\medskip
\end{minipage}\hspace{-0.2cm}
\begin{minipage}[b]{0.11\linewidth}
  \centering{\epsfig{figure=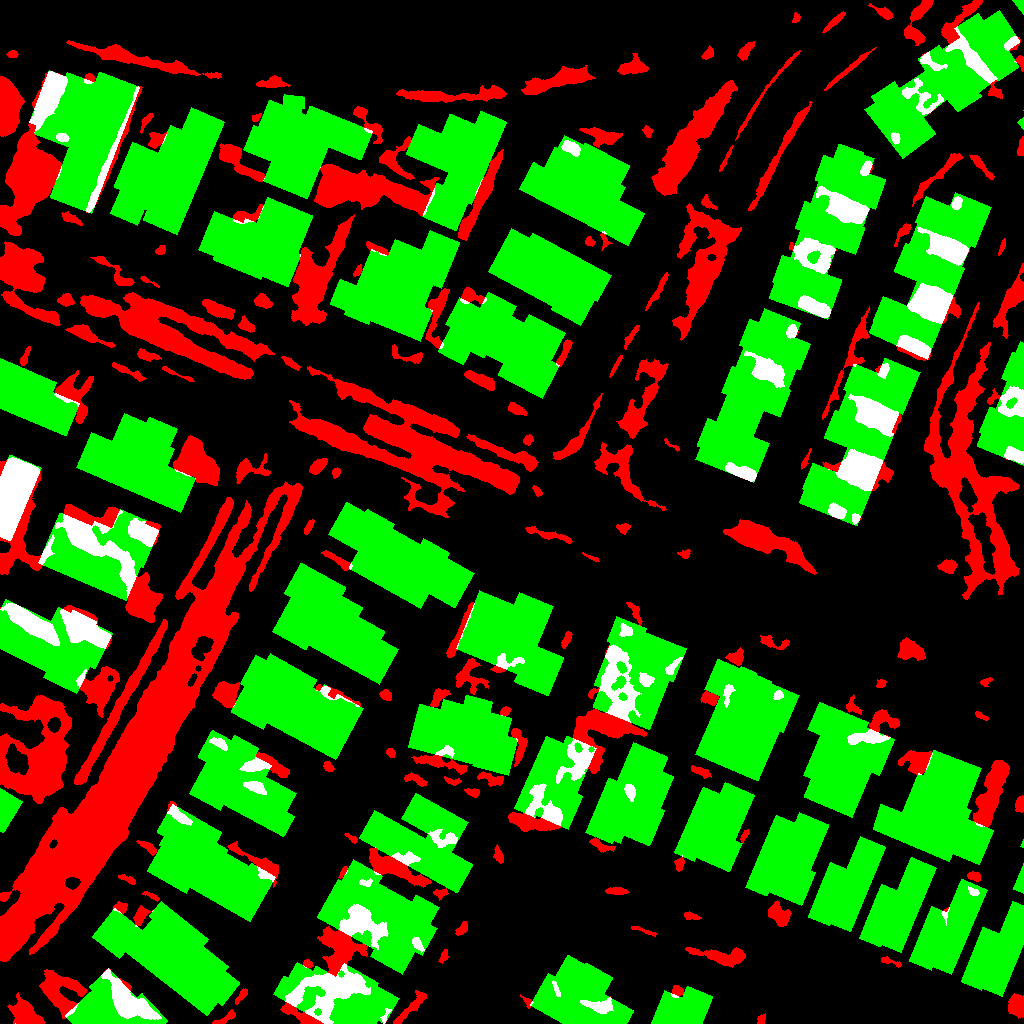,width=1.7cm}} 
  \centerline{(h)}\medskip
\end{minipage}\hspace{-0.2cm}
\begin{minipage}[b]{0.11\linewidth}
  \centering{\epsfig{figure=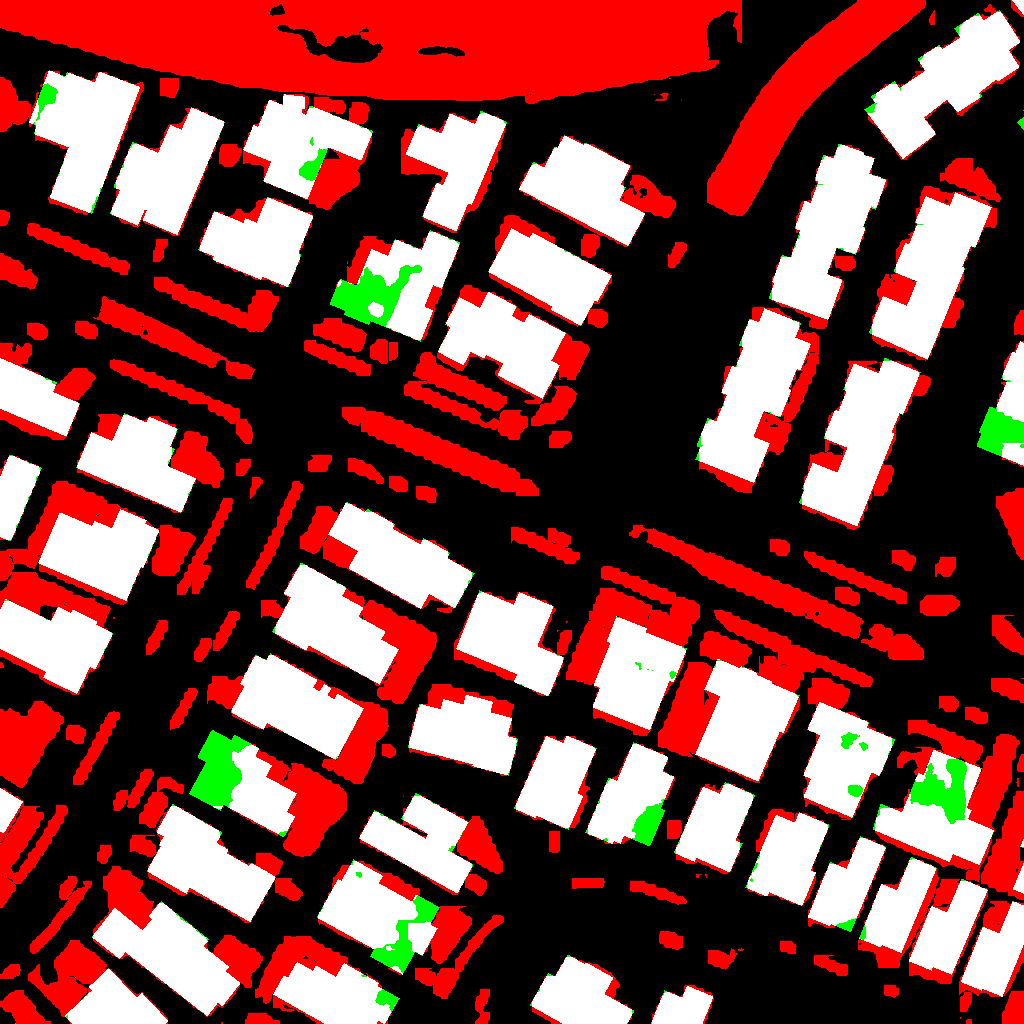,width=1.7cm}} 
  \centerline{(i)}\medskip
\end{minipage}\hspace{-0.2cm}
\vspace{-0.5cm}
\caption{Visualization results of different methods on WHU-CD dataset. (a) Image $X_1$. (b) Image $X_2$. (c) Ground Truth. (d) PCA-KM. (e) CNN-CD. (f) DSFA. (g) DCVA. (h) GMCD. (i) SCM (Ours). In change maps, white, black, red, and green represent TP, TN, FP, and FN, respectively.} 
\label{fig_res}
\end{figure*}

\textbf{Datasets}: We conducted external comparisons with other UCD methods and ablation study on two public binary CD datasets: LEVIR-CD \cite{chen_spatial-temporal_2020} and WHU-CD \cite{ji_fully_2019}. LEVIR-CD dataset is a building CD dataset consisting of 637 samples, each with a resolution of $1024\times1024\times3$. We directly experimented on the test set of LEVIR-CD, which contains 128 image pairs. In WHU-CD dataset, we equidistantly cut the whole image pair with the resolution of $32507 \times 15354$ in sliding-window manner, generating 480 test samples of $1024 \times 1024$. \\
\textbf{Evaluation Metrics}: We adopted three metrics to evaluate model performance, including F1 score, mean Intersection over Union (mIoU) score, and overall accuracy (OA). \\
\textbf{Compared Methods}: We chose five UCD methods for comparison. Traditional UCD approach includes PCA-KM \cite{turgay_celik_unsupervised_2009}, and DL-based approaches include CNN-CD \cite{jiang_convolutional_2016}, DSFA \cite{du_unsupervised_2019}, DCVA \cite{saha_unsupervised_2019} and GMCD \cite{tang_unsupervised_2022}. We replicated the aforementioned methods through open source codes and tested them on two datasets. \\
\textbf{Experimental Results}: Table.\ref{outer_exp_results} presents the performances of different UCD methods, our proposed SCM achieves the highest F1, mIoU and OA metrics on both datasets. We visualize two sets of predictions from different methods in Fig.\ref{fig_res}, where SCM achieves superior performances to other methods. Our method can segment related building changes under partial-change and thorough-change circumstances, with less pseudo changes and missed detections. \\
\textbf{Ablation Study}: We further performed ablation study on both datasets to demonstrate the effectiveness of the proposed recalibrated feature fusion (RFF) module and piecewise semantic attention (PSA) scheme. As shown in Table.\ref{ablation_study}, results show that the two components of the SCM crucially improve the accuracy of UCD task.

\begin{table}[htbp]
  \setlength{\abovedisplayskip}{4pt}
  \centering
  \caption{Performances of different UCD methods on LEVIR-CD and WHU-CD datasets}
  \begin{tabular}{p{1.8cm}<{\centering}p{0.61cm}<{\centering}p{0.61cm}<{\centering}p{0.61cm}<{\centering}p{0.61cm}<{\centering}p{0.61cm}<{\centering}p{0.61cm}<{\centering}}
    \toprule
    \multirow{2}[2]{*}{Method} & \multicolumn{3}{c}{LEVIR-CD} & \multicolumn{3}{c}{WHU-CD} \cr
    \cmidrule(lr){2-4} \cmidrule(lr){5-7}
     & F1 & mIoU & OA & F1 & mIoU & OA \\
    \midrule
    PCA-KM\cite{turgay_celik_unsupervised_2009} & 39.22 & 28.92 & 54.13 & 41.85 & 31.73 & 59.04 \\
    CNN-CD\cite{jiang_convolutional_2016} & 45.49 & 35.41 & 64.78 & 43.13 & 33.70 & 63.20 \\
    DSFA\cite{du_unsupervised_2019} & 47.65 & 40.70 & 77.33 & 48.39 & 42.46 & 81.02 \\
    DCVA\cite{saha_unsupervised_2019} & 52.89 & 46.09 & 84.75 & 53.90 & 44.89 & 79.44 \\
    GMCD\cite{tang_unsupervised_2022} & 50.02 & 42.75 & 79.64 & 55.55 & 47.56 & 84.49 \\
    \midrule
    SCM (Ours)& \textbf{62.80} & \textbf{53.67} & \textbf{88.80} & \textbf{61.81} & \textbf{52.14} & \textbf{86.14} \\
    \bottomrule
  \end{tabular}
  \label{outer_exp_results}
\end{table}%

\begin{table}[htbp]
  \centering
  \caption{Ablation Study of the proposed SCM}
  \begin{tabular}{p{1.6cm}<{\centering}p{0.61cm}<{\centering}p{0.61cm}<{\centering}p{0.61cm}<{\centering}p{0.61cm}<{\centering}p{0.61cm}<{\centering}p{0.61cm}<{\centering}}
    \toprule
    \multirow{2}[2]{*}{Method} & \multicolumn{3}{c}{LEVIR-CD} & \multicolumn{3}{c}{WHU-CD} \cr
    \cmidrule(lr){2-4} \cmidrule(lr){5-7}
     & F1 & mIoU & OA & F1 & mIoU & OA \\
    \midrule
    base & 42.21 & 30.12 & 53.35 & 42.24 & 30.61 & 54.87 \\
    base+RFF & 52.59 & 41.80 & 72.40 & 52.17 & 42.13 & 74.16 \\
    SCM & \textbf{62.80} & \textbf{53.67} & \textbf{88.80} & \textbf{61.81} & \textbf{52.14} & \textbf{86.14} \\
    \bottomrule
  \end{tabular}%
  \label{ablation_study}
\end{table}%

\section{Conclusion}
\label{sec_conclusion}

In this work, we introduced the Segment Change Model (SCM) for unsupervised change detection in VHR remote sensing images. A recalibrated feature fusion (RFF) module is proposed to integrate features and restore their semantic correlations across different scales. Cooperating with SAM and CLIP, we developed a piecewise semantic attention (PSA) scheme to further reduce pseudo change phenomenon. Experimental results affirm that our SCM outperforms conventional UCD methods. Nonetheless, the effect of varying semantic change targets on UCD performance requires further exploration.

\section{Acknowledgement}
\label{sec_ack}

This research was funded by the National Natural Science Foundation of China (No.42101346), the China Postdoctoral Science Foundation (No.2020M680109), and the Wuhan East Lake High-tech Development Zone Program of Unveiling and Commanding (No.2023KJB212). The numerical calculations were performed on the supercomputing system in the Supercomputing Center of Wuhan University.




\bibliographystyle{IEEEbib}
\bibliography{refs}

\end{document}